\documentclass[preprint,12pt,authoryear]{elsarticle}

\usepackage{lastpage}
\usepackage{bm}
\usepackage{amssymb}
\usepackage{amsmath}
\usepackage{amsfonts}
\usepackage{booktabs}
\usepackage{longtable}
\usepackage{setspace}
\usepackage{array}
\usepackage{autobreak}
\usepackage{graphicx}
\usepackage{subfigure}
\usepackage{float}
\usepackage[graphicx]{realboxes}
\usepackage{algorithm,algorithmic}
\usepackage{multirow}
\biboptions{sort&compress}
\usepackage{caption}
\usepackage[font=small,labelfont=bf,labelsep=none]{caption}
\journal{}
\usepackage{flushend}
\usepackage[T1]{fontenc}
\usepackage[colorlinks,bookmarksopen,bookmarksnumbered,citecolor=blue, linkcolor=blue, urlcolor=blue,breaklinks]{hyperref}
\usepackage{amsthm}
\theoremstyle{definition}
\newtheorem{example}{Example}
\begin{document}
	%\definecolor{ccr}{RGB}{10,110,150}
	\captionsetup[figure]{labelfont={bf},labelformat={default},labelsep=period,name={Fig.}}
	\begin{frontmatter}
		\title{Sparse Optimization for Transfer Learning: A $L_0$-Regularized Framework for Multi-Source Domain Adaptation}
		%\title{Sparsity-Driven Transfer Learning: A $L_0$-Regularized Framework for Multi-Source Domain Adaptation}
		%\title{Sparse Optimization for Transfer Learning in Heterogeneous Multi-Source Environments}
		%\title{A sparse optimization algorithm for transfer learning based on $L_0$ regularization}
		\author[1]{Chenqi Gong}
		\ead{gcq@stu.cqu.edu.cn}
		\author[1]{Hu Yang\corref{cor1}}
		\ead{yh@cqu.edu.cn}
		\cortext[cor1]{Corresponding author}
		\address[1]{College of Mathematics and Statistics, Chongqing University, Chongqing, 401331, China}
		\begin{abstract}
			This paper explores transfer learning in heterogeneous multi-source environments with distributional divergence between target and auxiliary domains. To address challenges in statistical bias and computational efficiency, we propose a Sparse Optimization for Transfer Learning (SOTL) framework based on $L_0$-regularization. 
			The method extends the Joint Estimation Transferred from Strata (JETS) paradigm with two key innovations: (1) $L_0$-constrained exact sparsity for parameter space compression and complexity reduction, and (2) refining optimization focus to emphasize target parameters over redundant ones. Simulations show that SOTL significantly improves both estimation accuracy and computational speed, especially under adversarial auxiliary domain conditions. Empirical validation on the Community and Crime benchmarks demonstrates the statistical robustness of the SOTL method in cross-domain transfer.
		\end{abstract}
		\begin{keyword}
			Transfer learning \sep $L_0$ regularization \sep High-dimension \sep Sparse linear model
		\end{keyword}
	\end{frontmatter}
	
	%\twocolumn
	%%%%%%%%%%%%%%%%%%%%%%%%%%%%%%%%%%%%%%%%%%%%%%
	%% Please use \tableofcontents for articles %%
	%% with 50 pages and more                   %%
	%%%%%%%%%%%%%%%%%%%%%%%%%%%%%%%%%%%%%%%%%%%%%%
	%\tableofcontents
	%\clearpage
\section{Introduction}\label{sec:1}
Modern statistical research increasingly involves datasets characterized by multi-source heterogeneity and diversity. Harnessing cross-domain data correlations to integrate multi-source information effectively and derive robust predictive inferences represents a critical challenge in contemporary machine learning. Transfer learning \citep{Torrey:2010} addresses this challenge by leveraging knowledge from related source domains to enhance target task performance through auxiliary information transfer. This paradigm has demonstrated significant success across diverse applications, including natural language processing \citep{Devlin:2019, Dettmers:2023}, text and data classification \citep{Pan:2010, Liu:2020, Cai:2021}, recommendation systems \citep{Pan:2013}, medical diagnosis \citep{Hajiramezanali:2018}, and image recognition and classification\citep{Oquab:2014, He:2020, Wang:2020, Choudhary:2023}.
	
Recent advancements have extended transfer learning to high-dimensional regimes. \citet{Bastani:2020} proposed a two-step transfer learning framework under single-source conditions, establishing estimation error bounds under $L_0$-sparse discrepancy between target and source parameters. \citet{Li:2021} developed the Trans Lasso algorithm for multi-source high-dimensional linear regression, while \citet{Tian:2022} extended this work to generalized linear models (Trans-GLM). \citet{Gao:2023} introduced Joint Estimation Transferred from Strata (JETS), a robust method for high-dimensional inference under limited samples and noisy auxiliary data. Despite these advances, persistent limitations remain: model parameter proliferation increases computational complexity, penalty parameter estimation exhibits covariate normalization dependency $\bm{X}$, and estimation bias persists under non-sparse regimes. These challenges have renewed interest in $L_0$ regularization for exact sparse recovery \citep{Huang:2018, Dai:2023, Ming:2024}. Inspired by the aforementioned research, we propose a Sparse Optimization for Transfer Learning (SOTL) framework based on $L_0$ regularization. The proposed method extends the Joint Estimation Transferred from Strata (JETS) paradigm through two key innovations:  
(1) exact sparsity enforcement: $L_0$-constrained variable selection to eliminate redundant parameters and reduce optimization complexity;  
(2) refining the optimization focus:  leveraging the computational properties of $L_0$ regularization to ensure emphasis on target parameters rather than irrelevant ones.
%Building upon the JETS approach, this approach incorporates $L_0$ penalty for variable selection, reducing the number of parameters to be optimized to lower computational costs, refining the estimation direction to enhance accuracy, and significantly accelerating the algorithm's runtime.

The remainder of the paper is structured as follows. Section \ref{sec:2} provides a detailed overview of the SOTL algorithm, outlining its methodology. Sections \ref{sec:3} and \ref{sec:4} present simulations and applications. A summary is provided in Section \ref{sec:5}.

\section{SOTL algorithm}\label{sec:2}
	We first define some symbols as follows for convenience. For the vector $\bm{\beta}=(\beta_1,\ldots,\beta_p)'$, denote its number of nonzero elements by $||\bm{\beta}||_0$ and define its $L_1$-norm by $||\bm{\beta}||_1$. 
	
	Assuming there are $Z$ groups of data sources, each group having $n_z$ observations and $p$-dimensional features where $z\in\lbrace 1,\ldots,Z\rbrace$ and $\sum\limits_{z=1}^Z n_z=N$. For the target data group of interest, let $n_t$ represent its number of observations, $\bm{y}^{(t)}\in\mathbb{R}^{n_t}$ denote its response variable, and $X^{(t)}\in\mathbb{R}^{n_t\times p}$ represent its design matrix. We focus on the target model:
	\begin{equation}
		\label{eq:1}
		\bm{y}^{(t)}=\bm{X}^{(t)}\bm{\beta}^{(t)}+\bm{\xi}^{(t)},
	\end{equation}
	where $\bm{\beta}^{(t)}\in \mathbb{R}^p$ is the parameter vector to be estimated, and $\bm{\xi}^{(t)} \in \mathbb{R}^{n_t} $ is a vector of random noise. Additionally, we suppose $p$ may substantially exceed $n_t$, $\bm{\beta}^{(t)}$ is sparse. For group $z\in\lbrace 1,\ldots,Z\rbrace$ with $z\ne t$, we focus on the model: 
	\begin{equation}
		\label{eq:2}
		\bm{y}^{(z)}=\bm{X}^{(z)}\bm{\beta}^{(z)}+\bm{\xi}^{(z)},
	\end{equation}
	where $\bm{y}^{(z)}\in\mathbb{R}^{n_z}$ is the response vector, $\bm{X}^{(z)}\in\mathbb{R}^{n_z\times p}$ is the design matrix, and $\bm{\xi}^{(z)} \in \mathbb{R}^{n_z}$ is the vector of random noise. 
	
	To address the potential correlation between the $Z-1$ auxiliary models and the target model (\ref{eq:1}), we introduce a weighting parameter $\bm{\omega}^{(z)}$ to quantify the deviation between $\bm{\beta}^{(z)}$ and $\bm{\beta}^{(t)}$, i.e., $\bm{\beta}^{(z)}=\bm{\beta}^{(t)}+\bm{\omega}^{(z)}$. Our objective is to utilize these $Z-1$ auxiliary models to estimate the target model. 
	
	\cite{Gao:2023} introduce the following model and its two variants, referred to as the JETS method, that is, 
	\begin{align}
		\label{eq:3}
		%\begin{split}
		(\hat{\bm{\beta}}^{(t)},\hat{\bm{\omega}}^{(z)})=&\mathop{\arg\min}\limits_{(\bm{\beta}^{(t)},\omega^{(z)})}\Big \{ \frac{1}{n_t}||\bm{y}^{(t)}-\bm{X}^{(t)}\bm{\beta}^{(t)}||^2_2\nonumber\\
		+&\sum\limits_{z=1,z\ne t}^{Z}\frac{1}{n_z}||\bm{y}^{(z)}-\bm{X}^{(z)}(\bm{\beta}^{(t)}+\bm{\omega}^{(z)}||^2_2\nonumber\\
		+&\lambda_t||\bm{\beta}^{(t)}||_1+\sum\limits_{z=1,z\ne t}^{Z}\lambda_z||\bm{\omega}^{(z)}||_1 \Big \}.
		%\end{split}
	\end{align}
	However, the model (\ref{eq:3}) involves two distinct parameters, making their optimization both complex and time-consuming. 
	
	To simplify the model, we set $\lambda_t=\lambda_z=\lambda$. Given data sets $\lbrace\bm{y}^{(z)},\bm{X}^{(z)}\rbrace$, we define artificial data sets $\mathbb{X}\in\mathbb{R}^{N\times pZ}$ and $\mathbb{Y}\in\mathbb{R}^{N}$ with sample size $N$ and dimension $p$, and define $\bm{\Phi}\in\mathbb{R}^{pZ\times 1}$ as follows:
	\[
	\scriptsize{
		\setlength{\arraycolsep}{1.0pt}
		\mathbb{X}=
		\begin{pmatrix}
			\bm{X}^{(t)}/\sqrt{n_t} & \bm{0} & \cdots & \bm{0} & \bm{0} & \cdots & \bm{0}\\
			
			\bm{X}^{(1)}/\sqrt{n_1} & \bm{X}^{(1)}/(\sqrt{n_1}) & \cdots & \bm{0} & \bm{0} & \cdots & \bm{0}\\
			
			\vdots & \vdots & \ddots &    &    &   & \bm{0}\\
			
			\bm{X}^{(t-1)}/\sqrt{n_{t-1}} & \bm{0} &   & \bm{X}^{(t-1)}/(\sqrt{n_{t-1}}) & \bm{0} &   & \bm{0}\\
			
			\bm{X}^{(t+1)}/\sqrt{n_{t+1}} & \bm{0} &   & \bm{0} & \bm{X}^{(t+1)}/(\sqrt{n_{t+1}}) &   & \bm{0}\\
			
			\vdots & \vdots &   &   &    & \ddots & \vdots\\
			
			\bm{X}^{(Z)}/\sqrt{n_Z} & \bm{0} & \cdots & \bm{0} & \bm{0} & \cdots & \bm{X}^{(Z)}/(\sqrt{n_Z})
	\end{pmatrix}},
	\]
	\[
	\small{
		\mathbb{Y}=
		\begin{pmatrix}
			\bm{y}^{(t)}/\sqrt{n_t}\\
			\bm{y}^{(1)}/\sqrt{n_1}\\
			\vdots\\
			\bm{y}^{(t-1)}/\sqrt{n_{t-1}}\\
			\bm{y}^{(t+1)}/\sqrt{n_{t+1}}\\
			\vdots\\
			\bm{y}^{(Z)}/\sqrt{n_Z}
		\end{pmatrix},
		\qquad
		\bm{\Phi}=
		\begin{pmatrix}
			\bm{\beta}^{(t)}\\
			\omega_1\\
			\vdots\\
			\omega_{(t-1)}\\
			\omega_{(t+1)}\\
			\vdots\\
			\omega_Z
	\end{pmatrix}}.
	\]
	The objective function to be minimized in (\ref{eq:3}) is transformed into the following problem, which we refer to as the S-JETS method for simplicity.
	\begin{equation}
		\label{eq:4}
		\hat{\bm{\Phi}}=\mathop{\arg\min}\limits_{\bm{\Phi}}\Big\{||\mathbb{Y}-\mathbb{X}\bm{\Phi}||_2^2+\lambda||\bm{\Phi}||_1\Big\}.
	\end{equation}
	
	Due to the bias introduced by the $L_1$-norm and the difficulty in determining the optimization focus between $\bm{\beta}$ and $\bm{\omega}$, we further incorporate the $L_0$-norm into the following model and refer to it as SOTL,
	\begin{equation}
		\label{eq:5}
		\hat{\bm{\Phi}}=\mathop{\arg\min}\limits_{\bm{\Phi}}\Big\{||\mathbb{Y}-\mathbb{X}\bm{\Phi}||_2^2+\lambda||\bm{\Phi}||_0\Big\}.
	\end{equation}
	The introduction of the $L_0$-norm eliminates regularization on the $\mathbb{X}$ matrix, thereby simplifying model estimation and reducing computational time. To determine the optimal size of important variables, decision-making can be guided by various information criteria, including the Bayesian Information Criterion (BIC) \citep{Schwarz:1978}, the Extended Bayesian Information Criterion (EBIC) \citep{Chen:2008}, and the Hierarchical Bayesian Information Criterion (HBIC) \citep{Wang:2013}. In this paper, we consider the following information criteria:
	\begin{equation}
		HBIC(\hat{\gamma})=Q(\hat{\bm{\Phi}})+\frac{C_N\log{(pZ)}}{N}||\hat{\bm{\Phi}}||_0,
	\end{equation}
	where $\hat{\gamma}=||\hat{\bm{\Phi}}||_0$ represents the sparse level, $Q(\hat{\bm{\Phi}})=\log{(\frac{1}{N}||\mathbb{Y}-\mathbb{X}\hat{\bm{\Phi}}||_2^2)}$ and $C_N=\log{(\log{(N)})}$. Finally, we summarize the SOTL algorithm in Algorithm \ref{alg:1}.
	\begin{algorithm}[!h]
		\caption{SOTL Algorithm}
		\label{alg:1}
		\renewcommand{\algorithmicrequire}{\textbf{Input:}}
		\renewcommand{\algorithmicensure}{\textbf{Output:}}
		\begin{algorithmic}[1]
			\REQUIRE artificial data sets $\lbrace\mathbb{Y},\mathbb{X}\rbrace$, with sample size $N$ and dimension $pZ$, maximum size of important variables $\Gamma$.  %%input
			\ENSURE the estimated coefficient vector $\hat{\bm{\beta}}(\hat{\gamma}_{opt})$.    %%output
			\FOR {$\gamma=1,2,\ldots,\Gamma$}
			\STATE Compute $\hat{\bm{\Phi}}(\gamma)$ using the R package abess \citep{Zhu:2022}, $||\hat{\bm{\Phi}}(\gamma)||_0=\gamma$;
			\STATE Calculate $HBIC(\gamma)$;
			\STATE Obtain the optimal size $\hat{\gamma}_{opt}=\arg\min(HBIC(\gamma))$;
			\ENDFOR
			\STATE Take the first $p$ dimensions of $\hat{\bm{\Phi}}(\hat{\gamma}_{opt})$, i.e., $\hat{\bm{\beta}}(\hat{\gamma}_{opt})$;
			\RETURN $\hat{\bm{\beta}}(\hat{\gamma}_{opt})$.
		\end{algorithmic}
	\end{algorithm}
	
	\section{Simulation studies}\label{sec:3}
	We demonstrate the performance of the SOTL method and compare it with the S-JETS method, as well as three variants of the JETS method: JETS-M1, JETS-M2, and JETS-M3. The R package abess \citep{Zhu:2022} was used to run SOTL, while the R package glmnet \citep{Friedman:2020} was used for S-JETS, JETS-M1, JETS-M2, and JETS-M3. The experiment was repeated $r$ times, and the following performance metrics were utilized: mean square error (MSE), support recovery accuracy (SRA), the number of nonzero coefficients (NZ), the false positive rate (FPR), the true positive rate (TPR), and the average running time (ART).
	\begin{itemize}
		\item $MSE=\frac{1}{r}\sum\limits_{i=1}^r{SE^{[i]}}$;
		\item $SRA=\frac{1}{r}\sum\limits_{i=1}^r\Big(\frac{|j\in\lbrace 1, \ldots, p\rbrace: \;\hat{\bm{\beta}}_{t,j}\ne 0\;\cap\;\bm{\beta}_{t,j}\ne 0\;+\;\hat{\bm{\beta}}_{t,j}=0\;\cap\;\bm{\beta}_{t,j}=0|}{|j\in\lbrace 1, \ldots, p\rbrace: \;\bm{\beta}_{t,j}|}\Big)^{[i]}$;
		\item $NZ=\frac{1}{r}\sum\limits_{i=1}^r{\Big(|j\in\lbrace 1, \ldots, p\rbrace: \;\hat{\bm{\beta}}_{t,j}\ne 0|\Big)}^{[i]}$;
		\item $FPR=\frac{1}{r}\sum\limits_{i=1}^r{\Big(\frac{|j\in\lbrace 1, \ldots, p\rbrace: \;\hat{\bm{\beta}}_{t,j}\ne 0\;\cap\;\bm{\beta}_{t,j}=0|}{|j\in\lbrace 1, \ldots, p\rbrace:\; \bm{\beta}_{t,j}=0|}\Big)}^{[i]}$;
		\item $TPR=\frac{1}{r}\sum\limits_{i=1}^r{\Big(\frac{|j\in\lbrace 1, \ldots, p\rbrace: \;\hat{\bm{\beta}}_{t,j}\ne 0\;\cap\;\bm{\beta}_{t,j}\ne0|}{|j\in\lbrace 1, \ldots, p\rbrace: \;\bm{\beta}_{t,j}\ne0|}\Big)}^{[i]}$;
		\item $ART=\frac{1}{r}\sum\limits_{i=1}^r{RT^{[i]}}$,
	\end{itemize}
	where $[i]$ represents the result of the i-th experiment, and $SE$ and $RT$ represents the prediction error and the runing time of a single experiment, respectively. In the simulation study, we aim for the MSE, FPR, and ART of the model to be as close to 0 as possible, while the SRA and TPR should be as close to 1 as possible. Additionally, the NZ should closely match the true number of non-zero coefficients.
	
	\subsection{Simulation settings}
	We consider three examples: Example \ref{ex:1}, where a single informative auxiliary group exists; Example \ref{ex:2}, where several informative auxiliary groups are incorporated; and Example \ref{ex:3}, where both informative and adversarial auxiliary groups coexist. For all three examples, we set the dimensionality $p=600$, the sparsity level $s=10$, and repeat the experiments $r=50$ times. Due to the time-consuming parameter updates in the methods of the three variants of JETS, in the expanded Example \ref{ex:2} and Example \ref{ex:3}, we only compare the SOTL method with the S-JETS method.
	\begin{example}
		\label{ex:1}
		In this example, we set the number of groups $Z=2$ and the target data group $t=1$. The target data design matrix $\bm{X}^{(1)}$ and the source data design matrix $\bm{X}^{(2)}$ are sampled from $N_p(0, \Sigma)$, $\Sigma_{j_1,j_2}=0.5^{|j_2-j_1|}$ for $j_1, j_2 = 1, \ldots, p$. We set $n_t=[60,70,80]$, $n_z=3n_t$, and set
		\begin{align}\nonumber
			\bm{\beta}^{(1)}=(\overbrace{2,\ldots,2}^{10},\overbrace{0,\ldots,0}^{590}),\;%\\\nonumber
			\bm{\omega}^{(1)}=(\overbrace{w,\ldots,w}^{10},\overbrace{0,\ldots,0}^{590}),\;%\\[3mm]\nonumber
			\bm{\beta}^{(2)}=\bm{\beta}^{(1)}+\bm{\omega}^{(1)},
		\end{align}
		where $w$ varies in $[0,0.5,1.0]$. Given $\bm{\beta}^{(1)}$, $\bm{\beta}^{(2)}$, $\bm{X}^{(1)}$ and $\bm{X}^{(2)}$, the vectors $\bm{Y}^{(1)}$ and $\bm{Y}^{(2)}$ can be obtained through the following expressions:
		$$
		\bm{Y}^{(1)}=\bm{X}^{(1)}\bm{\beta}^{(1)}+\bm{\xi}^{(1)},\quad \bm{Y}^{(2)}=\bm{X}^{(2)}\bm{\beta}^{(2)}+\bm{\xi}^{(2)},
		$$
		where each component of $\bm{\xi}^{(1)}$ and $\bm{\xi}^{(2)}$ is derived from a normal distribution $N(0,\sigma^2)$, where $\sigma$ varies in [0.2,0.5,0.8].
	\end{example}
	
	\begin{example}
		\label{ex:2} 
		In this example, we set the number of groups $Z=3$ and the target data group $t=1$. Let $n_t=[30,40,50]$, $n_{z_1}=n_{z_2}=n_t$. $\bm{X}^{(1)}$, $\bm{X}^{(2)}$ and $\bm{X}^{(3)}$ are generated as Example \ref{ex:1}. Let
	{\footnotesize
	\begin{align}
		\bm{\beta}^{(1)}=(\overbrace{2,\ldots,2}^{10},\overbrace{0,\ldots,0}^{590}),\;%\nonumber\\
		\bm{\omega}^{(1)}=(\overbrace{0.5w,\ldots,0.5w}^{10},\overbrace{0,\ldots,0}^{590}),\;%\nonumber\\
		\bm{\omega}^{(2)}=(\overbrace{w,\ldots,w}^{10},\overbrace{0,\ldots,0}^{590}),\nonumber
	\end{align}}
	where $w$ varies in $[0.2,0.4,0.6]$. Given $\bm{\beta}^{(1)}$, $\bm{\beta}^{(2)}$, $\bm{\beta}^{(3)}$, $\bm{X}^{(1)}$, $\bm{X}^{(2)}$ and $\bm{X}^{(3)}$, $\bm{Y}^{(1)}$, $\bm{Y}^{(2)}$ and $\bm{Y}^{(3)}$ can be obtained through the following expressions:
	\begin{align}
		\bm{Y}^{(1)}=\bm{X}^{(1)}\bm{\beta}^{(1)}+\bm{\xi}^{(1)},\;%\nonumber\\
		\bm{Y}^{(2)}=\bm{X}^{(2)}\bm{\beta}^{(2)}+\bm{\xi}^{(2)},\;%\nonumber\\
		\bm{Y}^{(3)}=\bm{X}^{(3)}\bm{\beta}^{(3)}+\bm{\xi}^{(3)},\nonumber
	\end{align}
	where each component of $\bm{\xi}^{(1)}$, $\bm{\xi}^{(2)}$ and $\bm{\xi}^{(3)}$ is derived from a normal distribution $N(0,\sigma^2)$, where $\sigma$ varies in $[2,2.5,3]$.
\end{example}
	\begin{example}
		\label{ex:3} In this example, we set the number of groups $Z=3$, the target data group $t=1$, $n_t=[40,60,80,100]$, and $n_{z_1}=n_{z_2}=n_t$. Set
		\begin{align}
			\bm{\beta}^{(1)}=(\overbrace{2,\ldots,2}^{10},\overbrace{0,\ldots,0}^{590}),\;%\nonumber\\
			\bm{\omega}^{(1)}=(\overbrace{0.5,\ldots,0.5}^{10},\overbrace{0,\ldots,0}^{590}),\;%\nonumber\\[3mm]
			\bm{\omega}_j^{(2)}=-4\nVdash_{\lbrace j\in J\rbrace},\nonumber
		\end{align}
		where $J$ is a randomly selected subset of $\lbrace 1,\ldots,p\rbrace$ and $|J|=40$. Each component of $\bm{\xi}^{(1)}$, $\bm{\xi}^{(2)}$ and $\bm{\xi}^{(3)}$ is derived from a normal distribution $N(0,\sigma^2)$, where $\sigma$ varies in $[0.5,1,1.5]$. All other settings remain consistent with those of Example \ref{ex:2}.
	\end{example}
	
	\subsection{Simulation results and analysis}
	Tables \ref{tab:1}-\ref{tab:3} present the results for Example \ref{ex:1}. As shown, these tables indicate that as the parameters \(\sigma\) and \(\omega\) increase, the MSE for all five methods tends to rise. This suggests that the increasing data variance and the divergence between the target and auxiliary datasets make the transfer models more challenging, resulting in reduced prediction accuracy. However, the SOTL method significantly enhances algorithm efficiency while maintaining estimation accuracy, demonstrating stable and superior performance across various predictive metrics compared to the other four methods.
	
	Table \ref{tab:4} summarizes the results for Example \ref{ex:2}, and Table \ref{tab:5} presents the results for Example \ref{ex:3}. Table \ref{tab:4} shows the results of model transfer estimation using two sets of information-based auxiliary data, each with a small sample size. In this example, the SOTL method outperforms the S-JETS method in nearly all six evaluation metrics, highlighting the superiority of SOTL in scenarios with small samples and high variance. Table \ref{tab:5} illustrates the estimation performance of the target model when the auxiliary dataset includes both informative and adversarial data. It is evident that the inclusion of adversarial auxiliary data can mislead model estimation, adversely affecting the results. However, under such conditions, the SOTL method demonstrates superior stability and speed compared to the S-JETS method, which suffers a significant decline in prediction accuracy. The SOTL method clearly outperforms the S-JETS method in the presence of unfavorable data sources.
	\renewcommand{\floatpagefraction}{.9}
	\begin{table}[!htbp]
		\centering
		\caption{Simulation results for Example \ref{ex:1} ($\sigma=0.2$).}
		\label{tab:1}
		\resizebox{0.85\linewidth}{!}{
			\begin{tabular}{ccccccccc}
				\toprule
				$\omega$ & Size  & Method & MSE   & SRA   & NZ    & FPR   & TPR   & ART \\
				\midrule
				\multirow{15}[2]{*}{$\omega=0$} & \multirow{5}[1]{*}{$n=60$} & SOTL  & \textbf{0.0427} & \textbf{1.0000} & \textbf{10.00} & \textbf{0.0000} & \textbf{1.0000} & 0.8044 \\
				&       & S-JETS & 0.1227 & 1.0000 & 10.00 & 0.0000 & 1.0000 & 0.3838 \\
				&       & JETS-M1 & 0.0655 & 0.9998 & 10.10 & 0.0002 & 1.0000 & 2.0278 \\
				&       & JETS-M2 & 0.0655 & 0.9998 & 10.10 & 0.0002 & 1.0000 & 3.8624 \\
				&       & JETS-M3 & 0.0654 & 1.0000 & 10.00 & 0.0000 & 1.0000 & 3.6162 \\
				& \multirow{5}[0]{*}{$n=70$} & SOTL  & \textbf{0.0419} & \textbf{1.0000} & \textbf{10.00} & \textbf{0.0000} & \textbf{1.0000} & 0.9194 \\
				&       & S-JETS & 0.1184 & 1.0000 & 10.00 & 0.0000 & 1.0000 & 0.4266 \\
				&       & JETS-M1 & 0.0569 & 0.9999 & 10.06 & 0.0001 & 1.0000 & 2.1728 \\
				&       & JETS-M2 & 0.0569 & 0.9999 & 10.06 & 0.0001 & 1.0000 & 4.1652 \\
				&       & JETS-M3 & 0.0567 & 1.0000 & 10.00 & 0.0000 & 1.0000 & 3.9240 \\
				& \multirow{5}[1]{*}{$n=80$} & SOTL  & \textbf{0.0418} & \textbf{1.0000} & \textbf{10.00} & \textbf{0.0000} & \textbf{1.0000} & 1.0334 \\
				&       & S-JETS & 0.1171 & 1.0000 & 10.00 & 0.0000 & 1.0000 & 0.4614 \\
				&       & JETS-M1 & 0.0589 & 1.0000 & 10.00 & 0.0000 & 1.0000 & 2.3696 \\
				&       & JETS-M2 & 0.0589 & 1.0000 & 10.00 & 0.0000 & 1.0000 & 4.5294 \\
				&       & JETS-M3 & 0.0598 & 1.0000 & 10.00 & 0.0000 & 1.0000 & 4.1866 \\
				\midrule
				\multirow{15}[1]{*}{$\omega=0.5$} & \multirow{5}[1]{*}{$n=60$} & SOTL  & \textbf{0.0484} & 0.9987 & 10.80 & 0.0014 & \textbf{1.0000} & 0.8056 \\
				&       & S-JETS & 0.0906 & 1.0000 & 10.00 & 0.0000 & 1.0000 & 0.4038 \\
				&       & JETS-M1 & 0.0501 & 0.9999 & 10.06 & 0.0001 & 1.0000 & 2.0684 \\
				&       & JETS-M2 & 0.1005 & 0.9990 & 10.60 & 0.0010 & 1.0000 & 3.9616 \\
				&       & JETS-M3 & 0.1004 & 1.0000 & 10.00 & 0.0000 & 1.0000 & 3.6900 \\
				& \multirow{5}[0]{*}{$n=70$} & SOTL  & \textbf{0.0459} & 0.9995 & 10.30 & 0.0005 & \textbf{1.0000} & 0.9232 \\
				&       & S-JETS & 0.0850 & 1.0000 & 10.00 & 0.0000 & 1.0000 & 0.4388 \\
				&       & JETS-M1 & 0.0475 & 1.0000 & 10.00 & 0.0000 & 1.0000 & 2.2546 \\
				&       & JETS-M2 & 0.0746 & 1.0000 & 10.02 & 0.0000 & 1.0000 & 4.3252 \\
				&       & JETS-M3 & 0.2094 & 1.0000 & 10.00 & 0.0000 & 1.0000 & 4.0344 \\
				& \multirow{5}[0]{*}{$n=80$} & SOTL  & \textbf{0.0459} & 0.9996 & 10.22 & 0.0004 & \textbf{1.0000} & 1.0132 \\
				&       & S-JETS & 0.0862 & 1.0000 & 10.00 & 0.0000 & 1.0000 & 0.4688 \\
				&       & JETS-M1 & 0.0477 & 1.0000 & 10.00 & 0.0000 & 1.0000 & 2.4098 \\
				&       & JETS-M2 & 0.0783 & 1.0000 & 10.02 & 0.0000 & 1.0000 & 4.5972 \\
				&       & JETS-M3 & 0.0812 & 1.0000 & 10.00 & 0.0000 & 1.0000 & 4.2824 \\
				\midrule
				\multirow{15}[1]{*}{$\omega=1$} & \multirow{5}[0]{*}{$n=60$} & SOTL  & \textbf{0.0489} & 0.9994 & 10.36 & 0.0006 & \textbf{1.0000} & 0.8146 \\
				&       & S-JETS & 0.1163 & 1.0000 & 10.00 & 0.0000 & 1.0000 & 0.3954 \\
				&       & JETS-M1 & 0.0524 & 1.0000 & 10.00 & 0.0000 & 1.0000 & 2.1324 \\
				&       & JETS-M2 & 0.0802 & 0.9999 & 10.08 & 0.0001 & 1.0000 & 4.1002 \\
				&       & JETS-M3 & 0.0893 & 1.0000 & 10.00 & 0.0000 & 1.0000 & 3.8130 \\
				& \multirow{5}[0]{*}{$n=70$} & SOTL  & \textbf{0.0466} & 0.9998 & 10.10 & 0.0002 & \textbf{1.0000} & 0.9260 \\
				&       & S-JETS & 0.1035 & 1.0000 & 10.00 & 0.0000 & 1.0000 & 0.4420 \\
				&       & JETS-M1 & 0.0497 & 1.0000 & 10.00 & 0.0000 & 1.0000 & 2.3278 \\
				&       & JETS-M2 & 0.0684 & 1.0000 & 10.02 & 0.0000 & 1.0000 & 4.4354 \\
				&       & JETS-M3 & 0.0711 & 1.0000 & 10.00 & 0.0000 & 1.0000 & 4.1120 \\
				& \multirow{5}[1]{*}{$n=80$} & SOTL  & \textbf{0.0457} & 0.9999 & 10.06 & 0.0001 & \textbf{1.0000} & 1.0192 \\
				&       & S-JETS & 0.0994 & 1.0000 & 10.00 & 0.0000 & 1.0000 & 0.4854 \\
				&       & JETS-M1 & 0.0471 & 1.0000 & 10.00 & 0.0000 & 1.0000 & 2.4818 \\
				&       & JETS-M2 & 0.0586 & 1.0000 & 10.00 & 0.0000 & 1.0000 & 4.7272 \\
				&       & JETS-M3 & 0.0601 & 1.0000 & 10.00 & 0.0000 & 1.0000 & 4.3652 \\
				\bottomrule
		\end{tabular}}
	\end{table}%
	
	\begin{table}[!htbp]
		\centering
		\caption{Simulation results for Example \ref{ex:1} ($\sigma=0.5$).}
		\label{tab:2}
		\resizebox{0.85\linewidth}{!}{
			\begin{tabular}{ccccccccc}
				\toprule
				$\omega$ & Size  & Method & MSE   & SRA   & NZ    & FPR   & TPR   & ART \\
				\midrule
				\multirow{15}[2]{*}{$\omega=0$} & \multirow{5}[1]{*}{$n=60$} & SOTL  & \textbf{0.2657} & \textbf{1.0000} & \textbf{10.00} & \textbf{0.0000} & \textbf{1} & \textbf{0.4090} \\
				&       & S-JETS & 0.3396 & 0.9965 & 12.10 & 0.0036 & 1     & 0.4374 \\
				&       & JETS-M1 & 0.3074 & 0.9904 & 15.76 & 0.0098 & 1     & 2.0144 \\
				&       & JETS-M2 & 0.3049 & 0.9877 & 17.40 & 0.0125 & 1     & 3.8580 \\
				&       & JETS-M3 & 0.2938 & 1.0000 & 10.00 & 0.0000 & 1     & 3.6204 \\
				& \multirow{5}[0]{*}{$n=70$} & SOTL  & \textbf{0.2605} & \textbf{1.0000} & \textbf{10.00} & \textbf{0.0000} & \textbf{1} & \textbf{0.4656} \\
				&       & S-JETS & 0.3278 & 0.9972 & 11.68 & 0.0028 & 1     & 0.4718 \\
				&       & JETS-M1 & 0.3010 & 0.9928 & 14.30 & 0.0073 & 1     & 2.1666 \\
				&       & JETS-M2 & 0.2982 & 0.9908 & 15.54 & 0.0094 & 1     & 4.1440 \\
				&       & JETS-M3 & 0.2812 & 1.0000 & 10.00 & 0.0000 & 1     & 3.8586 \\
				& \multirow{5}[1]{*}{$n=80$} & SOTL  & \textbf{0.2613} & \textbf{1.0000} & \textbf{10.00} & \textbf{0.0000} & \textbf{1} & 0.5298 \\
				&       & S-JETS & 0.3192 & 0.9979 & 11.26 & 0.0021 & 1     & 0.4968 \\
				&       & JETS-M1 & 0.2878 & 0.9793 & 22.44 & 0.0211 & 1     & 2.3268 \\
				&       & JETS-M2 & 0.2995 & 0.9788 & 22.72 & 0.0216 & 1     & 4.4316 \\
				&       & JETS-M3 & 0.2772 & 1.0000 & 10.02 & 0.0000 & 1     & 4.1364 \\
				\midrule
				\multirow{15}[1]{*}{$\omega=0.5$} & \multirow{5}[1]{*}{$n=60$} & SOTL  & \textbf{0.3274} & \textbf{1.0000} & \textbf{10.02} & \textbf{0.0000} & \textbf{1} & \textbf{0.4114} \\
				&       & S-JETS & 0.3331 & 0.9967 & 11.98 & 0.0034 & 1     & 0.4366 \\
				&       & JETS-M1 & 0.3093 & 0.9709 & 27.48 & 0.0296 & 1     & 2.0460 \\
				&       & JETS-M2 & 0.4195 & 0.9828 & 20.30 & 0.0175 & 1     & 3.9128 \\
				&       & JETS-M3 & 0.3810 & 1.0000 & 10.00 & 0.0000 & 1     & 3.6684 \\
				& \multirow{5}[0]{*}{$n=70$} & SOTL  & 0.3008 & 0.9999 & 10.06 & 0.0001 & \textbf{1} & \textbf{0.4664} \\
				&       & S-JETS & 0.3225 & 0.9982 & 11.10 & 0.0019 & 1     & 0.4880 \\
				&       & JETS-M1 & 0.2934 & 0.9983 & 11.02 & 0.0017 & 1     & 2.2012 \\
				&       & JETS-M2 & 0.4307 & 0.9972 & 11.70 & 0.0029 & 1     & 4.1926 \\
				&       & JETS-M3 & 0.4122 & 1.0000 & 10.00 & 0.0000 & 1     & 3.9014 \\
				& \multirow{5}[0]{*}{$n=80$} & SOTL  & 0.2933 & \textbf{1.0000} & 10.02 & \textbf{0.0000} & \textbf{1} & \textbf{0.5134} \\
				&       & S-JETS & 0.3101 & 0.9982 & 11.08 & 0.0018 & 1     & 0.5174 \\
				&       & JETS-M1 & 0.2931 & 0.9918 & 14.94 & 0.0084 & 1     & 2.3578 \\
				&       & JETS-M2 & 0.4652 & 1.0000 & 10.00 & 0.0000 & 1     & 4.4694 \\
				&       & JETS-M3 & 0.3836 & 1.0000 & 10.00 & 0.0000 & 1     & 4.1694 \\
				\midrule
				\multirow{15}[1]{*}{$\omega=1$} & \multirow{5}[0]{*}{$n=60$} & SOTL  & 0.3658 & \textbf{1.0000} & \textbf{10.00} & \textbf{0.0000} & \textbf{1} & \textbf{0.4134} \\
				&       & S-JETS & 0.3307 & 0.9975 & 11.52 & 0.0026 & 1     & 0.4388 \\
				&       & JETS-M1 & 0.3079 & 0.9841 & 19.54 & 0.0162 & 1     & 2.0932 \\
				&       & JETS-M2 & 0.4259 & 0.9789 & 22.68 & 0.0215 & 1     & 3.9756 \\
				&       & JETS-M3 & 0.6583 & 1.0000 & 10.00 & 0.0000 & 1     & 3.7102 \\
				& \multirow{5}[0]{*}{$n=70$} & SOTL  & 0.3207 & \textbf{1.0000} & 10.02 & \textbf{0.0000} & \textbf{1} & \textbf{0.4650} \\
				&       & S-JETS & 0.3278 & 0.9977 & 11.38 & 0.0023 & 1     & 0.4832 \\
				&       & JETS-M1 & 0.3013 & 0.9959 & 12.44 & 0.0041 & 1     & 2.2420 \\
				&       & JETS-M2 & 0.4059 & 0.9799 & 22.04 & 0.0204 & 1     & 4.2638 \\
				&       & JETS-M3 & 0.3509 & 1.0000 & 10.00 & 0.0000 & 1     & 3.9508 \\
				& \multirow{5}[1]{*}{$n=80$} & SOTL  & 0.3082 & \textbf{1.0000} & \textbf{10.00} & \textbf{0.0000} & \textbf{1} & \textbf{0.5218} \\
				&       & S-JETS & 0.3054 & 0.9985 & 10.90 & 0.0015 & 1     & 0.5230 \\
				&       & JETS-M1 & 0.2947 & 0.9845 & 19.30 & 0.0158 & 1     & 2.4042 \\
				&       & JETS-M2 & 0.3523 & 0.9813 & 21.20 & 0.0190 & 1     & 4.5644 \\
				&       & JETS-M3 & 0.3774 & 1.0000 & 10.00 & 0.0000 & 1     & 4.2190 \\
				\bottomrule
		\end{tabular}}%
	\end{table}
	
	\begin{table}[!htbp]
		\centering
		\caption{Simulation results for Example \ref{ex:1} ($\sigma=0.8$).}
		\label{tab:3}
		\resizebox{0.87\linewidth}{!}{
			\begin{tabular}{ccccccccc}
				\toprule
				$\omega$ & Size  & Method & MSE   & SRA   & NZ    & FPR   & TPR   & ART \\
				\midrule
				\multirow{15}[2]{*}{$\omega=0$} & \multirow{5}[1]{*}{$n=60$} & SOTL  & \textbf{0.6733} & \textbf{1.0000} & \textbf{10.00} & \textbf{0.0000} & \textbf{1.0000} & \textbf{0.4110} \\
				&       & S-JETS & 0.8538 & 0.9949 & 13.04 & 0.0052 & 1.0000 & 0.4456 \\
				&       & JETS-M1 & 0.7893 & 0.9028 & 68.34 & 0.0989 & 1.0000 & 2.0250 \\
				&       & JETS-M2 & 0.7893 & 0.9028 & 68.32 & 0.0988 & 1.0000 & 3.8910 \\
				&       & JETS-M3 & 0.6923 & 0.9976 & 11.46 & 0.0025 & 1.0000 & 3.6384 \\
				& \multirow{5}[0]{*}{$n=70$} & SOTL  & \textbf{0.6690} & \textbf{1.0000} & \textbf{10.00} & \textbf{0.0000} & \textbf{1.0000} & \textbf{0.4722} \\
				&       & S-JETS & 0.8315 & 0.9971 & 11.76 & 0.0030 & 1.0000 & 0.4804 \\
				&       & JETS-M1 & 0.7352 & 0.9625 & 32.52 & 0.0382 & 1.0000 & 2.2048 \\
				&       & JETS-M2 & 0.7352 & 0.9625 & 32.52 & 0.0382 & 1.0000 & 4.2006 \\
				&       & JETS-M3 & 0.7149 & 0.9999 & 10.06 & 0.0001 & 1.0000 & 3.9134 \\
				& \multirow{5}[1]{*}{$n=80$} & SOTL  & \textbf{0.6680} & \textbf{1.0000} & \textbf{10.00} & \textbf{0.0000} & \textbf{1.0000} & \textbf{0.5190} \\
				&       & S-JETS & 0.8003 & 0.9959 & 12.44 & 0.0041 & 1.0000 & 0.5238 \\
				&       & JETS-M1 & 0.7388 & 0.9860 & 18.42 & 0.0143 & 1.0000 & 2.3580 \\
				&       & JETS-M2 & 0.7388 & 0.9860 & 18.40 & 0.0142 & 1.0000 & 4.4692 \\
				&       & JETS-M3 & 0.6886 & 0.9999 & 10.04 & 0.0001 & 1.0000 & 4.1392 \\
				\midrule
				\multirow{15}[1]{*}{$\omega=0.5$} & \multirow{5}[1]{*}{$n=60$} & SOTL  & 0.8607 & \textbf{1.0000} & \textbf{10.00} & \textbf{0.0000} & \textbf{1.0000} & \textbf{0.4156} \\
				&       & S-JETS & 0.8283 & 0.9962 & 12.26 & 0.0038 & 1.0000 & 0.4488 \\
				&       & JETS-M1 & 0.7558 & 1.0000 & 10.00 & 0.0000 & 1.0000 & 2.0622 \\
				&       & JETS-M2 & 0.8823 & 0.9964 & 12.14 & 0.0036 & 1.0000 & 3.9756 \\
				&       & JETS-M3 & 1.0734 & 1.0000 & 10.00 & 0.0000 & 1.0000 & 3.8866 \\
				& \multirow{5}[0]{*}{$n=70$} & SOTL  & 0.7868 & \textbf{1.0000} & \textbf{10.00} & \textbf{0.0000} & \textbf{1.0000} & \textbf{0.4766} \\
				&       & S-JETS & 0.8115 & 0.9957 & 12.58 & 0.0044 & 1.0000 & 0.4948 \\
				&       & JETS-M1 & 0.7764 & 1.0000 & 10.00 & 0.0000 & 1.0000 & 2.3070 \\
				&       & JETS-M2 & 52.6043 & 0.9952 & 7.10  & 0.0000 & 0.7100 & 4.3958 \\
				&       & JETS-M3 & 11.8269 & 0.9994 & 9.64  & 0.0000 & 0.9640 & 4.0922 \\
				& \multirow{5}[0]{*}{$n=80$} & SOTL  & \textbf{0.7699} & \textbf{1.0000} & \textbf{10.02} & \textbf{0.0000} & \textbf{1.0000} & \textbf{0.5240} \\
				&       & S-JETS & 0.7758 & 0.9966 & 12.04 & 0.0035 & 1.0000 & 0.5296 \\
				&       & JETS-M1 & 0.8140 & 0.9688 & 28.74 & 0.0318 & 1.0000 & 2.4500 \\
				&       & JETS-M2 & 0.9415 & 0.9158 & 60.54 & 0.0857 & 1.0000 & 4.6744 \\
				&       & JETS-M3 & 0.8526 & 0.9999 & 10.04 & 0.0001 & 1.0000 & 4.3600 \\
				\midrule
				\multirow{15}[1]{*}{$\omega=1$} & \multirow{5}[0]{*}{$n=60$} & SOTL  & 0.7967 & \textbf{1.0000} & 10.02 & \textbf{0.0000} & \textbf{1.0000} & \textbf{0.4196} \\
				&       & S-JETS & 0.8409 & 0.9959 & 12.44 & 0.0041 & 1.0000 & 0.4542 \\
				&       & JETS-M1 & 0.7844 & 0.9994 & 10.36 & 0.0006 & 1.0000 & 2.1458 \\
				&       & JETS-M2 & 1.3559 & 1.0000 & 10.00 & 0.0000 & 1.0000 & 4.1068 \\
				&       & JETS-M3 & 2.3342 & 1.0000 & 10.00 & 0.0000 & 1.0000 & 3.8164 \\
				& \multirow{5}[0]{*}{$n=70$} & SOTL  & 0.8139 & \textbf{1.0000} & \textbf{10.00} & \textbf{0.0000} & \textbf{1.0000} & \textbf{0.4678} \\
				&       & S-JETS & 0.7964 & 0.9959 & 12.44 & 0.0041 & 1.0000 & 0.4958 \\
				&       & JETS-M1 & 0.7572 & 0.9996 & 10.24 & 0.0004 & 1.0000 & 2.3168 \\
				&       & JETS-M2 & 1.0432 & 0.9972 & 11.68 & 0.0028 & 1.0000 & 4.4234 \\
				&       & JETS-M3 & 0.9306 & 1.0000 & 10.00 & 0.0000 & 1.0000 & 4.0974 \\
				& \multirow{5}[1]{*}{$n=80$} & SOTL  & 0.7545 & \textbf{1.0000} & \textbf{10.00} & \textbf{0.0000} & \textbf{1.0000} & \textbf{0.5208} \\
				&       & S-JETS & 0.7859 & 0.9964 & 12.16 & 0.0037 & 1.0000 & 0.5380 \\
				&       & JETS-M1 & 0.7439 & 0.9861 & 18.34 & 0.0141 & 1.0000 & 2.5066 \\
				&       & JETS-M2 & 1.6604 & 1.0000 & 10.00 & 0.0000 & 1.0000 & 4.7422 \\
				&       & JETS-M3 & 100.8553 & 0.9839 & 0.34  & 0.0000 & 0.0340 & 4.3810 \\
				\bottomrule
		\end{tabular}}
	\end{table}%
	
	\begin{table}[!htbp]
		\centering
		\caption{Simulation results for Example \ref{ex:2}.}
		\label{tab:4}
		\resizebox{0.78\linewidth}{!}{
			\begin{tabular}{cccccccccc}
				\toprule
				$\sigma$ & $\omega$ & Size  & Method & MSE   & SRA   & NZ    & FPR   & TPR   & ART \\
				\midrule
				\multirow{18}[5]{*}{$\sigma=2$} & \multirow{6}[2]{*}{$\omega=0.2$} & \multirow{2}[1]{*}{$n=30$} & SOTL  & \textbf{4.7940} & \textbf{0.9998} & \textbf{10.06} & \textbf{0.0001} & 0.9980 & \textbf{0.3600} \\
				&       &       & S-JETS & 7.1086 & 0.9937 & 13.78 & 0.0064 & 1.0000 & 0.3964 \\
				&       & \multirow{2}[0]{*}{$n=40$} & SOTL  & \textbf{4.5378} & \textbf{1.0000} & \textbf{10.02} & \textbf{0.0000} & \textbf{1.0000} & \textbf{0.3980} \\
				&       &       & S-JETS & 6.0002 & 0.9966 & 12.04 & 0.0035 & 1.0000 & 0.4458 \\
				&       & \multirow{2}[1]{*}{$n=50$} & SOTL  & \textbf{4.3832} & \textbf{1.0000} & \textbf{10.00} & \textbf{0.0000} & \textbf{1.0000} & \textbf{0.4360} \\
				&       &       & S-JETS & 5.4310 & 0.9970 & 11.82 & 0.0031 & 1.0000 & 0.5066 \\
				\cmidrule{2-10}
				& \multirow{6}[2]{*}{$\omega=0.4$} & \multirow{2}[1]{*}{$n=30$} & SOTL  & \textbf{5.2260} & \textbf{0.9996} & \textbf{10.02} & \textbf{0.0002} & 0.9880 & \textbf{0.3560} \\
				&       &       & S-JETS & 6.9929 & 0.9936 & 13.86 & 0.0065 & 1.0000 & 0.4042 \\
				&       & \multirow{2}[0]{*}{$n=40$} & SOTL  & \textbf{4.8889} & \textbf{0.9999} & \textbf{9.98} & \textbf{0.0000} & 0.9960 & \textbf{0.3886} \\
				&       &       & S-JETS & 5.7576 & 0.9961 & 12.32 & 0.0039 & 1.0000 & 0.4448 \\
				&       & \multirow{2}[1]{*}{$n=50$} & SOTL  & \textbf{4.7129} & \textbf{1.0000} & \textbf{10.00} & \textbf{0.0000} & \textbf{1.0000} & \textbf{0.4380} \\
				&       &       & S-JETS & 5.4505 & 0.9968 & 11.92 & 0.0033 & 1.0000 & 0.5076 \\
				\cmidrule{2-10}          & \multirow{6}[1]{*}{$\omega=0.6$} & \multirow{2}[1]{*}{$n=30$} & SOTL  & \textbf{5.9780} & \textbf{0.9994} & \textbf{9.96} & \textbf{0.0003} & 0.9800 & \textbf{0.3690} \\
				&       &       & S-JETS & 6.3799 & 0.9954 & 12.78 & 0.0047 & 1.0000 & 0.3896 \\
				&       & \multirow{2}[0]{*}{$n=40$} & SOTL  & \textbf{5.3446} & \textbf{0.9999} & \textbf{10.04} & \textbf{0.0001} & 0.9980 & \textbf{0.3956} \\
				&       &       & S-JETS & 5.6643 & 0.9956 & 12.66 & 0.0045 & 1.0000 & 0.4412 \\
				&       & \multirow{2}[0]{*}{$n=50$} & SOTL  & \textbf{5.3168} & \textbf{1.0000} & \textbf{10.00} & \textbf{0.0000} & \textbf{1.0000} & \textbf{0.4416} \\
				&       &       & S-JETS & 5.3403 & 0.9946 & 13.24 & 0.0055 & 1.0000 & 0.4996 \\
				\midrule
				\multirow{18}[5]{*}{$\sigma=2.5$} & \multirow{6}[1]{*}{$\omega=0.2$} & \multirow{2}[0]{*}{$n=30$} & SOTL  & \textbf{8.0560} & \textbf{0.9988} & \textbf{10.26} & \textbf{0.0008} & 0.9780 & \textbf{0.3610} \\
				&       &       & S-JETS & 11.0978 & 0.9936 & 13.82 & 0.0065 & 0.9980 & 0.3976 \\
				&       & \multirow{2}[0]{*}{$n=40$} & SOTL  & \textbf{7.1817} & \textbf{0.9997} & \textbf{10.08} & \textbf{0.0002} & 0.9940 & \textbf{0.3958} \\
				&       &       & S-JETS & 9.3609 & 0.9963 & 12.22 & 0.0038 & 1.0000 & 0.4362 \\
				&       & \multirow{2}[1]{*}{$n=50$} & SOTL  & \textbf{6.9053} & \textbf{0.9999} & \textbf{10.08} & \textbf{0.0001} & \textbf{1.0000} & \textbf{0.4430} \\
				&       &       & S-JETS & 8.5866 & 0.9965 & 12.10 & 0.0036 & 1.0000 & 0.5082 \\
				\cmidrule{2-10}          & \multirow{6}[2]{*}{$\omega=0.4$} & \multirow{2}[1]{*}{$n=30$} & SOTL  & \textbf{8.6108} & \textbf{0.9991} & \textbf{10.08} & \textbf{0.0005} & 0.9760 & \textbf{0.3620} \\
				&       &       & S-JETS & 10.7902 & 0.9951 & 12.96 & 0.0050 & 1.0000 & 0.4128 \\
				&       & \multirow{2}[0]{*}{$n=40$} & SOTL  & \textbf{7.7459} & \textbf{0.9995} & \textbf{10.12} & \textbf{0.0004} & 0.9900 & \textbf{0.4000} \\
				&       &       & S-JETS & 9.1357 & 0.9957 & 12.56 & 0.0043 & 1.0000 & 0.4524 \\
				&       & \multirow{2}[1]{*}{$n=50$} & SOTL  & \textbf{7.3970} & \textbf{0.9998} & \textbf{9.96} & \textbf{0.0001} & 0.9920 & \textbf{0.4426} \\
				&       &       & S-JETS & 8.7406 & 0.9971 & 11.72 & 0.0029 & 1.0000 & 0.5222 \\
				\cmidrule{2-10}          & \multirow{6}[2]{*}{$\omega=0.6$} & \multirow{2}[1]{*}{$n=30$} & SOTL  & \textbf{8.4792} & \textbf{0.9992} & \textbf{9.86} & \textbf{0.0003} & 0.9680 & \textbf{0.3648} \\
				&       &       & S-JETS & 10.2374 & 0.9949 & 13.00 & 0.0051 & 0.9980 & 0.4058 \\
				&       & \multirow{2}[0]{*}{$n=40$} & SOTL  & \textbf{8.2986} & \textbf{0.9995} & \textbf{9.98} & \textbf{0.0002} & 0.9840 & \textbf{0.3986} \\
				&       &       & S-JETS & 9.0762 & 0.9952 & 12.88 & 0.0049 & 1.0000 & 0.4434 \\
				&       & \multirow{2}[1]{*}{$n=50$} & SOTL  & \textbf{7.6781} & \textbf{0.9999} & \textbf{9.96} & \textbf{0.0000} & 0.9960 & \textbf{0.4436} \\
				&       &       & S-JETS & 8.1321 & 0.9965 & 12.10 & 0.0036 & 1.0000 & 0.5168 \\
				\midrule
				\multirow{18}[6]{*}{$\sigma=3$} & \multirow{6}[2]{*}{$\omega=0.2$} & \multirow{2}[1]{*}{$n=30$} & SOTL  & \textbf{14.7981} & \textbf{0.9968} & \textbf{8.98} & \textbf{0.0007} & 0.8540 & \textbf{0.3668} \\
				&       &       & S-JETS & 16.5013 & 0.9930 & 14.14 & 0.0071 & 0.9960 & 0.4128 \\
				&       & \multirow{2}[0]{*}{$n=40$} & SOTL  & \textbf{11.6304} & \textbf{0.9987} & \textbf{9.86} & \textbf{0.0005} & 0.9540 & \textbf{0.4022} \\
				&       &       & S-JETS & 14.4491 & 0.9955 & 12.68 & 0.0045 & 1.0000 & 0.4548 \\
				&       & \multirow{2}[1]{*}{$n=50$} & SOTL  & \textbf{10.0837} & \textbf{0.9996} & \textbf{10.06} & \textbf{0.0002} & 0.9920 & \textbf{0.4482} \\
				&       &       & S-JETS & 12.7538 & 0.9961 & 12.34 & 0.0040 & 1.0000 & 0.5200 \\
				\cmidrule{2-10}          & \multirow{6}[2]{*}{$\omega=0.4$} & \multirow{2}[1]{*}{$n=30$} & SOTL  & \textbf{14.6548} & \textbf{0.9971} & \textbf{9.32} & \textbf{0.0009} & 0.8780 & \textbf{0.3682} \\
				&       &       & S-JETS & 15.4295 & 0.9947 & 13.16 & 0.0054 & 0.9980 & 0.4118 \\
				&       & \multirow{2}[0]{*}{$n=40$} & SOTL  & \textbf{11.4938} & \textbf{0.9989} & \textbf{10.00} & \textbf{0.0005} & 0.9680 & \textbf{0.3982} \\
				&       &       & S-JETS & 13.4497 & 0.9957 & 12.58 & 0.0044 & 1.0000 & 0.4438 \\
				&       & \multirow{2}[1]{*}{$n=50$} & SOTL  & \textbf{10.6232} & \textbf{0.9995} & \textbf{9.98} & \textbf{0.0002} & 0.9840 & \textbf{0.4434} \\
				&       &       & S-JETS & 12.0132 & 0.9971 & 11.76 & 0.0030 & 1.0000 & 0.5196 \\
				\cmidrule{2-10}          & \multirow{6}[2]{*}{$\omega=0.6$} & \multirow{2}[1]{*}{$n=30$} & SOTL  & 14.9348 & \textbf{0.9969} & \textbf{9.32} & \textbf{0.0010} & 0.8740 & \textbf{0.3666} \\
				&       &       & S-JETS & 14.8705 & 0.9966 & 11.98 & 0.0034 & 0.9960 & 0.4002 \\
				&       & \multirow{2}[0]{*}{$n=40$} & SOTL  & \textbf{11.8995} & \textbf{0.9989} & \textbf{9.78} & \textbf{0.0004} & 0.9560 & \textbf{0.4042} \\
				&       &       & S-JETS & 12.7732 & 0.9964 & 12.16 & 0.0037 & 1.0000 & 0.4400 \\
				&       & \multirow{2}[1]{*}{$n=50$} & SOTL  & \textbf{11.3194} & \textbf{0.9995} & \textbf{9.96} & \textbf{0.0002} & 0.9820 & \textbf{0.4438} \\
				&       &       & S-JETS & 11.9256 & 0.9968 & 11.92 & 0.0033 & 1.0000 & 0.5198 \\
				\bottomrule
		\end{tabular}}
	\end{table}%
	
	\begin{table}[!htbp]
		\centering
		\caption{Simulation results for Example \ref{ex:3}.}
		\label{tab:5}%
		\resizebox{1\linewidth}{!}{
			\begin{tabular}{ccccccccc}
				\toprule
				$\sigma$ & Size  & Method & MSE   & SRA   & NZ    & FPR   & TPR   & ART \\
				\midrule
				\multirow{6}[2]{*}{$\sigma=0.5$} & \multirow{2}[1]{*}{$n=30$} & SOTL  & \textbf{26.3484} & \textbf{0.9906} & \textbf{5.26} & \textbf{0.0007} & \textbf{0.4820} & \textbf{0.3946} \\
				&       & S-JETS & 80.6440 & 0.9863 & 3.06  & 0.0011 & 0.2420 & 0.4142 \\
				& \multirow{2}[0]{*}{$n=40$} & SOTL  & \textbf{22.4562} & \textbf{0.9911} & \textbf{5.08} & \textbf{0.0004} & \textbf{0.4860} & \textbf{0.4402} \\
				&       & S-JETS & 64.0587 & 0.9895 & 4.66  & 0.0008 & 0.4180 & 0.4602 \\
				& \multirow{2}[1]{*}{$n=50$} & SOTL  & \textbf{19.8748} & 0.9919 & 5.14  & \textbf{0.0000} & 0.5140 & \textbf{0.4842} \\
				&       & S-JETS & 52.2817 & 0.9922 & 5.98  & 0.0006 & 0.5640 & 0.5298 \\
				\midrule
				\multirow{6}[2]{*}{$\sigma=1$} & \multirow{2}[1]{*}{$n=30$} & SOTL  & \textbf{31.1793} & \textbf{0.9897} & \textbf{5.30} & 0.0013 & \textbf{0.4560} & \textbf{0.3878} \\
				&       & S-JETS & 78.6001 & 0.9871 & 3.60  & 0.0011 & 0.2940 & 0.4114 \\
				& \multirow{2}[0]{*}{$n=40$} & SOTL  & \textbf{21.4311} & \textbf{0.9916} & 5.18  & \textbf{0.0002} & 0.5080 & \textbf{0.4328} \\
				&       & S-JETS & 54.2244 & 0.9915 & 5.86  & 0.0008 & 0.5380 & 0.4654 \\
				& \multirow{2}[1]{*}{$n=50$} & SOTL  & \textbf{22.6589} & 0.9912 & 4.90  & \textbf{0.0001} & 0.4820 & \textbf{0.4808} \\
				&       & S-JETS & 53.3042 & 0.9915 & 5.78  & 0.0007 & 0.5340 & 0.5436 \\
				\midrule
				\multirow{6}[2]{*}{$\sigma=1.5$} & \multirow{2}[1]{*}{$n=30$} & SOTL  & \textbf{29.5150} & \textbf{0.9897} & \textbf{5.24} & 0.0012 & \textbf{0.4520} & \textbf{0.3884} \\
				&       & S-JETS & 79.0725 & 0.9871 & 3.00  & 0.0006 & 0.2620 & 0.4094 \\
				& \multirow{2}[0]{*}{$n=40$} & SOTL  & \textbf{24.9257} & \textbf{0.9909} & \textbf{4.92} & \textbf{0.0003} & \textbf{0.4720} & \textbf{0.4396} \\
				&       & S-JETS & 69.9474 & 0.9886 & 4.30  & 0.0009 & 0.3740 & 0.4824 \\
				& \multirow{2}[1]{*}{$n=50$} & SOTL  & \textbf{24.1518} & \textbf{0.9911} & 4.92  & \textbf{0.0002} & \textbf{0.4800} & \textbf{0.4936} \\
				&       & S-JETS & 63.0483 & 0.9904 & 5.06  & 0.0007 & 0.4640 & 0.5612 \\
				\bottomrule
		\end{tabular}}
		\label{tab:addlabel}%
	\end{table}%
	
	\section{Empirical studies}\label{sec:4}
	\subsection{Experimental setting}
	The proposed algorithm was implemented on the Communities and Crime dataset, which can be accessed at \url{https://archive.ics.uci.edu/dataset/183/communities+and+crime}. This dataset consists of 99 dimensions for each data point. We conducted 100 repeated experiments, with each iteration involving random resampling of the target dataset to determine the samples used for the training and test sets. The data settings for the two experiments are as follows:\\
	\textbf{Experiment 1 (Single-source auxiliary group experiment)}: In this experiment, 322 samples were selected from the dataset. The target dataset consists of crime data from State 1, with 43 samples, of which 30 were designated as the training set and 13 as the test set. The auxiliary dataset includes crime data from State 6, with 278 samples, all assigned to the training set.\\
	\textbf{Experiment 2 (Multi-source auxiliary group experiment)}: In this experiment, 436 samples were selected from the dataset. The target dataset consists of crime data from State 9, comprising 69 samples, with 44 assigned to the training set and 25 to the test set. The auxiliary dataset includes crime data from State 34 and State 48, with 211 and 156 samples, respectively, all allocated to the training set.

	\subsection{Result analysis}
	\begin{figure*}[!htbp]
		\centering %????
		\includegraphics[height=9.1cm,width=13cm]{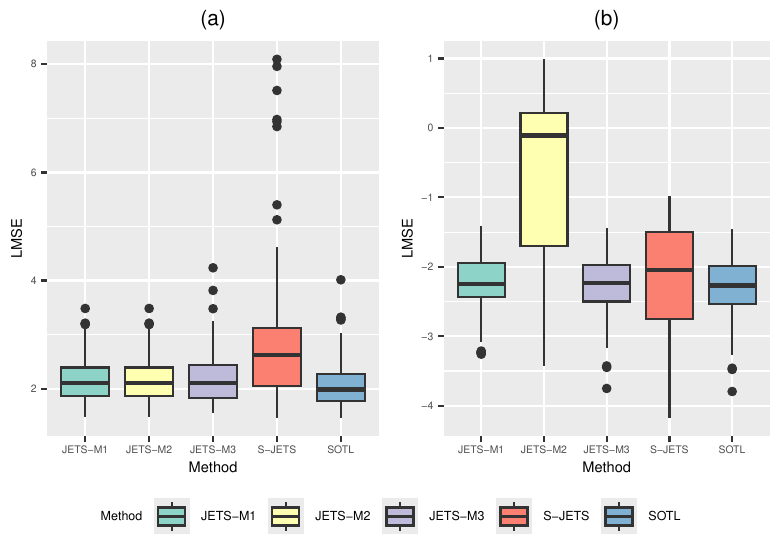}
		\caption{Comparison of mean square errors for five methods in Experiment 1 (a) and Experiment 2 (b).}
		\label{fig:1}
	\end{figure*}
	
	Figure \ref{fig:1} (a) shows the logarithm of the mean squared error (LMSE) for crime rate predictions across five methods, evaluated over 100 repeated experiments in Experiment 1. The results indicate that the SOTL method achieves the lowest mean squared error, reflecting superior predictive performance. The three variants of the JETS method—JETS-M1, JETS-M2, and JETS-M3—demonstrate moderate performance, while the S-JETS method exhibits the poorest performance on this dataset.
	
	Figure \ref{fig:1} (b) presents the logarithm of the mean squared error (LMSE) for crime rate predictions across five methods, based on 100 repeated experiments conducted in Experiment 2. The results clearly show that, with the inclusion of multi-source auxiliary data, the SOTL method maintains the highest predictive accuracy, while the JETS-M2 method displays instability. The mean squared errors for JETS-M1, JETS-M2, JETS-M3, and S-JETS are all higher than that of the SOTL algorithm.
	\section{Summary}\label{sec:5}
	In this paper, we propose a Sparse Optimization for Transfer Learning (SOTL) framework based on $L_0$ regularization. This framework improves the penalization term in JETS, overcoming the limitations of the $L_1$-norm and extending transfer learning to high-dimensional statistics. It demonstrates robust performance across various simulation scenarios, with significant improvements over JETS. In empirical studies, we apply the SOTL algorithm to the Community and Crime datasets, yielding promising results with superior performance compared to existing methods. Overall, SOTL reduces the number of tuning parameters, significantly accelerates computational runtime, and maintains high estimation accuracy, making it a strong candidate for real-world applications.
	
	\section*{Declaration of competing interest}
	The authors declare that they have no known competing financial interests or personal relationships that could have appeared to
	influence the work reported in this paper.
	
	\section*{Acknowledges}
	This work is supported by the National Natural Science Foundation of China [Grant No.12371281].
	
	\section*{Data availability}
	The data acquisition link is provided in the paper.
	%\newpage


\begin{thebibliography}{00}
		\bibitem[Torrey \& Shavlik(2010)]{Torrey:2010} Torrey, L., \& Shavlik, J. (2010) Transfer learning. In E. Soria, J. Martin, R. Magdalena, M. Martinez, A. Serrano (Eds.), \textit{Handbook of Research on Machine Learning Applications} (pp. 242-264). IGI Global.
		
		\bibitem[Devlin et al.(2019)]{Devlin:2019} Devlin, J., Chang, M., Lee, K., \& Toutanova, K. (2019). BERT: pre-training of deep bidirectional transformers for language understanding. In J. Burstein, C. Doran, T. Solorio (Eds.), \textit{Proceedings of the 2019 Conference of the North American Chapter of the Association for Computational Linguistics: Human Language Technologies, Volume 1 (Long and Short Papers)} (pp. 4171-4186). Association for Computational Linguistics. 
		
		\bibitem[Dettmers et al.(2023)]{Dettmers:2023} Dettmers, T., Pagnoni, A., Holtzman, A., \& Zettlemoyer, L. (2023). QLORA: efficient finetuning of quantized LLMs. In A. Oh, T. Naumann, A. Globerson, K. Saenko, M. Hardt, S. Levine (Eds.), \textit{Proceedings of the 37th International Conference on Neural Information Processing Systems} (pp. 10088-10115). Curran Associates, Inc.
		
		\bibitem[Pan \& Yang(2010)]{Pan:2010} Pan, S. J., \& Yang, Q. (2010). A survey on transfer learning. \textit{IEEE Transactions on Knowledge and Data Engineering}, 22(10),1345-1359. \url{https://doi.org/10.1109/TKDE.2009.191}.
		
		\bibitem[Liu et al.(2020)]{Liu:2020} Liu, Z., Li, X., Qiao, L., \& Durrani, D. K. (2020). A cross-region transfer learning method for classification of community service cases with small datasets. \textit{Knowledge-Based Systems}, 193, Article 105390. \url{https://doi.org/10.1016/j.knosys.2019.105390}.
		
		\bibitem[Cai \& Wei(2021)]{Cai:2021} Cai, T. T., \& Wei, H. (2021). Transfer learning for nonparametric classification: minimax rate and adaptive classifier. \textit{The Annals of Statistics}, 49(1), 100-128. \url{https://doi.org/10.1214/20-AOS1949}.
		
		\bibitem[Pan \& Yang(2013)]{Pan:2013} Pan, W., \& Yang, Q. (2013). Transfer learning in heterogeneous collaborative filtering domains. \textit{Artificial Intelligence}, 197, 39-55. \url{https://doi.org/10.1016/j.artint.2013.01.003}.
		
		\bibitem[Hajiramezanali et al.(2018)]{Hajiramezanali:2018} Hajiramezanali, E., Dadaneh, S. Z., Karbalayghareh, A., Zhou, M., \& Qian, X. (2018). Bayesian multi-domain learning for cancer subtype discovery from next-generation sequencing count data. In S. Bengio, H. Wallach, H. Larochelle, K. Grauman, N. Cesa-Bianchi, R. Garnett (Eds.), \textit{Proceedings of the 32nd International Conference on Neural Information Processing Systems} (pp. 9133-9142). Curran Associates Inc.
		
		\bibitem[Oquab et al.(2014)]{Oquab:2014} Oquab, M., Bottou, L., Laptev, I., \& Sivic, J. (2014). Learning and transferring mid-level image representations using convolutional neural networks. In \textit{Proceedings of the 2014 IEEE Conference on Computer Vision and Pattern Recognition} (pp. 1717-1724). IEEE Computer Society. %\url{https://doi.org/10.1109/CVPR.2014.222}
		
		\bibitem[He et al.(2020)]{He:2020} He, X., Chen, Y., \& Ghamisi, P. (2020). Heterogeneous transfer learning for hyperspectral image classification based on convolutional neural network. \textit{IEEE Transactions on Geoscience and Remote Sensing}, 58(5), 3246-3263. \url{https://doi.org/10.1109/TGRS.2019.2951445}.
		
		\bibitem[Wang et al.(2020)]{Wang:2020} Wang, S., Zhang, L., \& Fu, J. (2020). Adversarial transfer learning for cross-domain visual recognition. \textit{Knowledge-Based Systems}, 204, Article 106258. \url{https://doi.org/10.1016/j.knosys.2020.106258}.
		
		\bibitem[Choudhary et al.(2023)]{Choudhary:2023} Choudhary, T., Gujar, S., Goswami, A., Mishra, V., \& Badal, T. (2023). Deep learning-based important weights-only transfer learning approach for COVID-19 CT-scan classification. \textit{Applied Intelligence}, 53(6), 7201-7215. \url{https://doi.org/10.1007/s10489-022-03893-7}.
		
		\bibitem[Bastani(2020)]{Bastani:2020} Bastani, H. (2020). Predicting with proxies: transfer learning in high dimension. \textit{Management Science}, 67(5), 2657-3320. \url{https://doi.org/10.1287/mnsc.2020.3729}.
		
		\bibitem[Li et al.(2021)]{Li:2021} Li, S., Cai, T. T., \& Li, H. (2021). Transfer learning for high-dimensional linear regression: Prediction, estimation and minimax optimality. \textit{Journal of the Royal Statistical Society Series B: Statistical Methodology}, 84(1), 149-173. \url{https://doi.org/10.1111/rssb.12479}.
		
		\bibitem[Tian \& Feng(2022)]{Tian:2022} Tian, Y., \& Feng, Y. (2022). Transfer learning under high-dimensional generalized linear models. \textit{Journal of the American Statistical Association}, 118(544), 2684-2697. \url{https://doi.org/10.1080/01621459.2022.2071278}.
		
		\bibitem[Gao \& Yang(2023)]{Gao:2023} Gao, Y., \& Yang, Y. (2023). Transfer learning on stratified data: Joint estimation transferred from strata. \textit{Pattern Recognition}, 140, Article 109535. \url{https://doi.org/10.1016/j.patcog.2023.109535}.
		
		%\url{https://doi.org/}
		
		\bibitem[Huang et al.(2018)]{Huang:2018} Huang, J., Jiao, Y., Liu, Y., \& Lu, X. (2018). A constructive approach to $L_0$ penalized regression. \textit{Journal of Machine Learning Research}, 19(10), 1-37.
		
		\bibitem[Dai(2023)]{Dai:2023} Dai, S. (2023). Variable selection in convex quantile regression: $L_1$-norm or $L_0$-norm regularization? \textit{European Journal of Operational Research}, 305(1), 338-355. \url{https://doi.org/10.1016/j.ejor.2022.05.041}.
		
		\bibitem[Ming \& Yang(2024)]{Ming:2024} Ming, H., \& Yang, H. (2024). $L_0$ regularized logistic regression for large-scale data. \textit{Pattern Recognition}, 146, Article 110024. \url{https://doi.org/10.1016/j.patcog.2023.110024}.
		
		\bibitem[Schwarz(1978)]{Schwarz:1978} Schwarz, G. (1978). Estimating the dimension of a model. \textit{The Annals of Statistics}, 6(2), 461-464. \url{https://doi.org/10.1214/aos/1176344136}.
		
		\bibitem[Chen \& Chen(2008)]{Chen:2008} Chen, J., \& Chen, Z. (2008). Extended Bayesian information criteria for model selection with large model spaces. \textit{Biometrika}, 95(3), 759-771. \url{https://doi.org/10.1093/biomet/asn034}.
		
		\bibitem[Wang et al.(2013)]{Wang:2013} Wang, L., Kim, Y., \& Li, R. (2013). Calibrating non-convex penalized regression in ultra-high dimension. \textit{The Annals of Statistics}, 41(5), 2505-2536. \url{https://doi.org/10.1214/13-AOS1159}.
		
		\bibitem[Zhu et al.(2022)]{Zhu:2022} Zhu, J., Wang, X., Hu, L., Huang, J., Jiang, K., Zhang, Y., Lin, S., \& Zhu, J. (2022). Abess: a fast best-subset selection library in Python and R. \textit{Journal of Machine Learning Research}, 23(202), 1-7. %\url{https://doi.org/10.32614/cran.package.abess}
		
		\bibitem[Friedman et al.(2010)]{Friedman:2020} Friedman, J., Hastie, T., \& Tibshirani, R. (2010). Regularization paths for generalized linear models via coordinate descent. \textit{Journal of Statistical Software}, 33(1), 1-22. \url{https://doi.org/10.18637/jss.v033.i01}.
		
	\end{thebibliography}
\end{document}